\documentclass{article}

\usepackage{microtype}
\usepackage{color, colortbl}
\definecolor{Gray}{gray}{0.98}
\usepackage{graphicx}
\usepackage{subfigure}
\usepackage{pifont}
\newcommand{\cmark}{\ding{51}}%
\newcommand{\xmark}{\ding{55}}%
\usepackage{booktabs} 
\usepackage{hyperref}

\usepackage[accepted]{icml2020}
\usepackage{array}
\newcolumntype{d}[1]{D{.}{.}{#1}}
\newcolumntype{P}[1]{>{\centering\arraybackslash}p{#1}}
\usepackage{amsfonts}       
\usepackage{nicefrac}       
\usepackage{epsfig}
\usepackage{amsmath}
\usepackage{amssymb}
\usepackage{multirow}

\DeclareMathOperator*{\argmax}{arg\,max}
\DeclareMathOperator*{\argmin}{arg\,min}

\newcommand{\yy}{{\mathbf y}}

\newcommand{\ftheta}{f_{\theta}}
\newcommand{\xx}{\mathbf x}
\newcommand{\Ncal}{\mathcal{N}}
\newcommand{\Ecal}{\mathcal{E}}
\newcommand{\Bcal}{\mathcal{B}}
\newcommand{\Rcal}{\mathcal{R}}
\newcommand{\Lcal}{\mathcal{L}}
\newcommand{\LL}{\mathbf L}
\newcommand{\xxq}{\mathbf{x}}
\newcommand{\zz}{\mathbf{z}}
\newcommand{\XXb}{\mathbb{X}_{\text{base}}}
\newcommand{\XXs}{\mathbb{X}_{\text{s}}}
\newcommand{\XXq}{\mathbb{X}_{\text{q}}}
\newcommand{\YY}{\mathbf Y}
\newcommand{\DD}{\mathbf D}

\newcommand{\one}{{\mathbf 1}}
\newcommand{\WW}{\mathbf W}
\newcommand{\mm}{{\mathbf m}}
\newcommand{\II}{{\mathbf I}}
\usepackage{mathtools}

\newcommand{\ceq}{\stackrel{\mathclap{\normalfont\mbox{c}}}{=}}

\begin{document}

\twocolumn[
\icmltitle{Laplacian Regularized Few-Shot Learning}

\icmlsetsymbol{equal}{*}

\begin{icmlauthorlist}
\icmlauthor{Imtiaz Masud Ziko}{ets}
\icmlauthor{Jose Dolz}{ets}
\icmlauthor{Eric Granger}{ets}
\icmlauthor{Ismail~Ben~Ayed}{ets}
\end{icmlauthorlist}

\icmlaffiliation{ets}{\'ETS Montreal, Canada}
\icmlcorrespondingauthor{Imtiaz Masud Ziko}{imtiaz-masud.ziko.1@etsmtl.ca}

\icmlkeywords{Few-Shot Learning, Nearest Neighbor, Laplacian, Concave-Convex procedure}
\vskip 0.3in
]

\printAffiliationsAndNotice{}  

\begin{abstract}
We propose a transductive Laplacian-regularized inference for few-shot tasks. Given any feature embedding learned from the base classes, we minimize a quadratic binary-assignment function containing two terms: (1) a unary term assigning query samples to the nearest class prototype, and (2) a pairwise Laplacian term encouraging nearby query samples to have consistent label assignments. Our transductive inference does not re-train the base model, and can be viewed as a graph clustering of the query set, subject to supervision constraints from the support set. We derive a computationally efficient bound optimizer of a relaxation of our function, which computes independent (parallel) updates for each query sample, while guaranteeing convergence. Following a simple cross-entropy training on the base classes, and without complex meta-learning strategies, we conducted comprehensive experiments over five few-shot learning benchmarks. Our LaplacianShot consistently outperforms state-of-the-art methods by significant margins across different models, settings, and data sets. Furthermore, our transductive inference is very fast, with computational times that are close to inductive inference, and can be used for large-scale few-shot tasks. 

\end{abstract}

\section{Introduction}
\label{sec:introduction}

Deep learning models have achieved human-level performances in various tasks. The success of these models rely considerably on exhaustive learning from large-scale labeled data sets. 
Nevertheless, they still have difficulty generalizing to novel classes unseen during training, given only a few labeled instances for these new classes. In contrast, humans can 
learn new tasks easily from a handful of examples, by leveraging prior experience and related context. Few-shot learning \cite{fei2006one,Miller-Matsakis-Viola2000,Vinyals2016MatchingNF} 
has emerged as an appealing paradigm to bridge this gap. Under standard few-shot learning scenarios, a model is first trained on substantial labeled data over an initial set of classes, 
often referred to as the base classes. Then, supervision for novel classes, which are unseen during base training, is limited to just one or few labeled examples per class. The model 
is evaluated over few-shot tasks, each one supervised by a few labeled examples per novel class (the support set) and containing unlabeled samples for evaluation (the query set).

The problem has recently received substantial research interests, with a large body of work based on complex meta-learning and episodic-training strategies. The meta-learning setting uses the base training data to create a set of few-shot tasks (or episodes), with support and query samples that simulate generalization difficulties during test times, and train the model to generalize well on these artificial tasks. For example, \cite{Vinyals2016MatchingNF} introduced matching network, which employs an attention mechanism to predict the unknown query samples as a linear combination of the support labels, while using episodic training and memory architectures. Prototypical networks \cite{snell2017prototypical} maintain a single prototype representation for each class in the embedding space, and minimize the negative log-probability of the query features with episodic training. Ravi \& Larochelle \citeyearpar{Ravi2017OptimizationAA} viewed optimization as a model for few-shot learning, and used an LSTM meta-learner to update classifier parameters. Finn et al. \citeyearpar{Finn2017ModelAgnosticMF} proposed MAML, a meta-learning strategy that attempts to make a model ``easy'' to fine-tune. These widely adopted works were recently followed by an abundant meta-learning literature, for instance, \cite{sung2018learning,oreshkin2018tadam, mishra2018a, rusu2018metalearning, liu2018learning,can,ye2020fewshot}, among many others.

Several recent studies explored transductive inference for few-shot tasks, e.g., \cite{liu2018learning,can,Dhillon2020A,hu2020empirical,kim2019edge,team}, among others. Given a few-shot task at test time, transductive inference performs class predictions jointly for all the unlabeled query samples of the task, rather than one sample at a time as in inductive inference.
For instance, TPN \cite{liu2018learning} used label propagation \cite{z2004learning} along with episodic training and a specific network architecture, so as to learn how to propagate labels from labeled to unlabeled samples. CAN-T \cite{can} is another meta-learning based transductive method, which uses attention mechanisms to propagate labels to unlabeled query samples. The transductive fine-tuning method by \cite{Dhillon2020A} re-train the network by minimizing an additional entropy loss, which encourages peaked (confident) class predictions at unlabeled query points, in conjunction with a standard cross-entropy loss defined on the labeled support set. 

Transductive few-shot methods typically perform better than their inductive
counterparts. However, this may come at the price of a much heavier computational complexity during inference. For example, the entropy fine-tuning in \cite{Dhillon2020A} re-trains the network, performing gradient updates over all the parameters during inference. Also, the label propagation in \cite{liu2018learning} requires a matrix inversion, which has a computational overhead that is cubic with respect to the number of query samples. This may be an impediment for deployment for large-scale few-shot tasks.

We propose a transductive Laplacian-regularized inference for few-shot tasks. Given any feature embedding learned from the base data, our method minimizes a quadratic binary-assignment function integrating two 
types of potentials: (1) unary potentials assigning query samples to the nearest class prototype, and (2) pairwise potentials favoring consistent label assignments for nearby query samples. 
Our transductive inference can be viewed as a graph clustering of the query set, subject to supervision constraints from the support set, and does not re-train the base model. Following a relaxation of our function, we derive a computationally efficient bound optimizer, which computes independent (parallel) label-assignment updates for each query point, with guaranteed convergence. We conducted comprehensive experiments on five few-shot learning benchmarks, with different levels of difficulties. Using a simple cross-entropy training on the base classes, and without complex meta-learning strategies, our LaplacianShot outperforms state-of-the-art methods by significant margins, consistently providing improvements across different settings, data sets, and training models. Furthermore, our transductive inference is very fast, with computational times that are close to inductive inference, and can be used for large-scale tasks.

\section{Laplacian Regularized Few-Shot Learning}
\subsection{Proposed Formulation}
In the few-shot setting, we are given a labeled support set $\XXs=\bigcup_{c=1}^{C} \XXs^c$ with $C$ test classes, where each novel class $c$ has $|\XXs^c|$ labeled examples, for instance, $|\XXs^c|=1$ for 1-shot and $|\XXs^c|=5$ for 5-shot. The objective of few-shot learning is, therefore, to accurately classify unlabeled unseen query sample set $\XXq=\bigcup_{c=1}^{C} \XXq^c$ from these $C$ test classes. This setting is referred to as the $|\XXs^c|$-shot $C$-way few-shot learning.

Let $\ftheta$ denotes the embedding function of a deep convolutional neural network, with parameters $\theta$ and $\xx_q=\ftheta(\zz_q) \in \mathbb{R}^M$ encoding the features of a given data point $\zz_q$. Embedding $\ftheta$ is learned from a labeled training set $\XXb$, with  
base classes
that are different from 
the few-shot classes of $\XXs$ and $\XXq$.
In our work, parameters $\theta$ are learned through a basic network training with the standard cross-entropy loss defined over $\XXb$, without resorting to any complex episodic-training or meta-learning strategy. For each query feature point $\xxq_{q}$ in a few-shot task, we define a latent binary assignment vector $\yy_q = [y_{q,1}, \dots, y_{q,C}]^t \in\{0,1\}^C$, which is within the $C$-dimensional probability simplex $\nabla_C = \{\yy \in [0, 1]^C \; | \; {\mathbf 1}^t \yy = 1 \}$: binary $y_{q,c}$ is equal to 1 if $\xxq_q$ belongs to class $c$, and equal to 0 otherwise. $t$ is used as the transpose operator. Let $\YY$ denotes the $N \times C$ matrix whose rows are formed by $\yy_q$, where $N$ is the number of query points in $\XXq$. We propose a transductive few-shot inference, which minimizes a Laplacian-regularization objective for few-shot tasks w.r.t assignment variables $\YY$, subject to simplex and integer constraints $\yy_q \in \nabla_C$ and $\yy_q \in\{0,1\}^C$, $\forall q$:  
\begin{eqnarray}
\label{eq:LN}
{\Ecal}(\YY)  & = & {\Ncal(\YY)} + \frac{\lambda}{2} {\Lcal(\YY)} \\
{\Ncal(\YY)} & = & \sum_{q=1}^{N} \sum_{c=1}^C y_{q,c} d(\xxq_q - \mm_c) \nonumber\\
 {\Lcal(\YY)} & = & \frac{1}{2} \sum_{q,p} w (\xxq_q, \xxq_p) \|\yy_q - \yy_p\|^2 \nonumber 
\end{eqnarray}
In \eqref{eq:LN}, the first term $\Ncal(\YY)$ is minimized globally when each query point is assigned to the class of the nearest prototype $\mm_c$ from the support set, using a distance metric $d(\xxq_q,\mm_c)$, such as the Euclidean distance. In the \textbf{1-shot} setting, prototype $\mm_c$ is the support example of class c, whereas in \textbf{multi-shot}, $\mm_c$ can be the  mean of the support examples. In fact, $\mm_c$ can be further rectified 
by integrating information from the query features, as we will detail later in our experiments. 
\begin{algorithm}[tb]
  \caption{Proposed Algorithm for LaplacianShot}
  \label{alg}
\begin{algorithmic}
  \STATE {\bfseries Input:} $\XXs$, $\XXq$, $\lambda$, $\ftheta$
  \STATE {\bfseries Output:} $Labels \in \{1,..,C\}^{N}$ for $\XXq$
  \STATE Get prototypes $\mm_c$.
  \STATE Compute $\mathbf{a}_q$ using \eqref{aq}~$\forall \xxq_q \in \XXq$.
  \STATE Initialize $i = 1$.
  \STATE Initialize $\yy^i_{q} =\frac{\exp(-\mathbf{a}_q)}{\mathbf{1}^t\exp(-\mathbf{a}_q)}$.
  \REPEAT
  \STATE Compute $\yy^{i+1}_q$ using \eqref{closed-form}
  \STATE $\yy^{i}_q \leftarrow \yy^{i+1}_q$.
  \STATE $\YY = [\yy^i_q];\; \forall q$.
  \STATE $i = i+1$.
  \UNTIL{$\Bcal_i(\YY)$ in \eqref{Aux-function} does not change}
  \STATE $l_q = \underset{c}{\argmax}~\yy_q ;\; \forall \yy_q \in \YY$.
  \STATE $Labels = \{l_q\}_{q=1}^{N}$
\end{algorithmic}
\end{algorithm}

The second term $\Lcal(\YY)$ is the well-known Laplacian regularizer, which can be equivalently  written as $\mathrm{tr}(\YY^t \LL \YY)$, where $\LL$ is the Laplacian matrix\footnote{The Laplacian matrix corresponding 
to affinity matrix $\WW = [w (\xxq_q, \xxq_p)]$ is $\LL= \DD-\WW$, with $\DD$ the diagonal matrix whose diagonal elements are given by: $D_q = \sum_pw(\mathbf{x}_q,\mathbf{x}_p)$.} 
corresponding 
to affinity matrix $\WW = [w (\xxq_q, \xxq_p)]$, and $\mathrm{tr}$ denotes the trace operator. Pairwise potential $w (\xxq_q, \xxq_p)$ 
evaluates the similarity between feature vectors $\xxq_q$ and $\xxq_p$, and can be computed
using some kernel function. 
The Laplacian term encourages nearby points ($\xxq_q$, $\xxq_p$) in the feature space to have the same latent label assignment, thereby regularizing 
predictions at query samples for few-shot tasks. As we will show later in our comprehensive experiments, the pairwise Laplacian term complements the unary potentials in $\Ncal(\YY)$, substantially increasing the predictive performance of few-shot learning across different networks, and various benchmark datasets with different levels of difficulty. 

More generally, Laplacian regularization is widely used in the contexts of graph clustering \cite{VonLuxburg2007,ShiMalik2000,ziko2018scalable,WangCarreira-Perpinan2014} and semi-supervised learning \cite{weston2012deep, belkin2006manifold}. For instance, popular spectral graph clustering techniques \cite{VonLuxburg2007,ShiMalik2000} optimize the Laplacian term subject to partition-balance constraints. In this connection, our transductive inference can be viewed as a graph clustering of the query set, subject to supervision constraints from the support set. 

Regularization parameter $\lambda$ controls the trade-off between the two terms. It is worth noting that the recent nearest-prototype classification in \cite{wang2019simpleshot} corresponds to the particular case of $\lambda = 0$ of our model in \eqref{eq:LN}. It assigns a query sample $\xxq_q$ to the label of the closest support prototype in the feature space, thereby minimizing ${\Ncal(\YY)}$:
\begin{equation}
    y_{q,c^{*}} = 1\quad \text{if}~\quad c^{*} = \argmin_{c \in \{1,\ldots, C\}}d(\xxq_q,\mm_{c})
    \label{eq:initial_pred}
\end{equation}

\subsection{Optimization}
In this section, we propose an efficient bound-optimization technique for solving a relaxed version of our objective in \eqref{eq:LN}, which guarantees convergence, while computing independent closed-form updates for each query sample in few-shot tasks. It is well known that minimizing pairwise functions over binary variables is NP-hard \cite{Tian-AAAI}, and a standard approach in the context of clustering algorithms is to relax the integer constraints, for instance, using a convex \cite{WangCarreira-Perpinan2014} or a concave relaxation \cite{ziko2018scalable}. In fact, by relaxing integer constraints $\yy_q \in\{0,1\}^C$, our objective in \eqref{eq:LN} becomes a convex quadratic problem. However, this would require solving for the $N \times C$ assignment variables all together, with additional projections steps for handling the simplex constraints. In this work, we use a concave relaxation of the Laplacian-regularized objective in \eqref{eq:LN}, which, as we will later show, yields fast independent and closed-form updates for each assignment variable, with convergence guarantee. Furthermore, it enables us to draw interesting connections between Laplacian regularization and attention mechanisms in few-shot learning \cite{Vinyals2016MatchingNF}. 

It is easy to verify that, for binary (integer) simplex variables, the Laplacian term in \eqref{eq:LN} can be written as follows, after some simple manipulations: 
\begin{equation}
\label{tight-relaxation}
\Lcal(\YY) = \sum_{q} D_{q} - \sum_{q,p} w (\xxq_q, \xxq_p) \yy_{q}^{t} \yy_p
\end{equation} 
where $D_q = \sum_pw(\mathbf{x}_q,\mathbf{x}_p)$ denotes the {\em degree} of query sample $\mathbf{x}_q$. 
By relaxing integer constraints $\yy_q \in\{0,1\}^C$, the expression in Eq. \eqref{tight-relaxation} can be viewed as a {\em concave} relaxation\footnote{Equality \eqref{tight-relaxation} holds in for points on the vertices of the simplex, i.e.,  
$\yy_q \in\{0,1\}^C$, but is an approximation for points within the simplex (soft assignments), i.e., $\yy_q \in ]0,1[^C$.} for Laplacian term $\Lcal(\YY)$ when symmetric affinity matrix $\WW = [w (\xxq_q, \xxq_p)]$ is positive semi-definite. 
As we will see in the next paragraph, concavity is important to derive an efficient bound optimizer for our model, with independent and closed-form updates for each query sample. Notice that the first term in relaxation \eqref{tight-relaxation} is a constant independent of the soft (relaxed) assignment variables.  

We further augment relaxation \eqref{tight-relaxation} with a convex negative-entropy barrier function $\yy_q^{t} \log \yy_q$, 
which avoids expensive projection steps and Lagrangian-dual inner iterations for the simplex constraints of each query point. 
Such a barrier\footnote{Note that entropy-like barriers are known in the context of Bregman-proximal optimization \cite{Yuan2017}, and have well-known computational benefits when dealing with simplex constraints.} removes the need for extra dual variables for constraints $\yy_q \geq 0$ by 
restricting the domain of each assignment variable to non-negative values, and yields closed-form updates for the dual variables of 
constraints $\mathbf{1}^t \yy_q = 1$. Notice that this barrier function is null at the vertices of the simplex. Putting all together, and omitting the additive constant $\sum_{q} D_{q}$ in \eqref{tight-relaxation}, we minimize the following concave-convex relaxation of our objective in \eqref{eq:LN} w.r.t soft assignment variables $\YY$, subject to simplex constraints $\yy_q \in \nabla_C, \forall q$:  
\begin{equation}
\label{tight-relaxation-2}
{\Rcal}(\YY) = \YY^{t} \log \YY + {\Ncal(\YY)} + \frac{\lambda}{2} {\tilde{\Lcal}(\YY)}
\end{equation}
where $\tilde{\Lcal}(\YY) = - \sum_{q,p} w (\xxq_q, \xxq_p) \yy_{q}^{t} \yy_p$.  

{\bf Bound optimization:} In the following, we detail an iterative bound-optimization solution for relaxation \eqref{tight-relaxation-2}. Bound optimization, often referred to as MM (Majorize-Minimization) framework \cite{lange2000optimization, Zhang2007}, is a general optimization principle\footnote{The general MM principle is widely used in machine learning in various problems as it enables to replace a difficult optimization problem with a sequence of easier sub-problems \cite{Zhang2007}. Examples of well-known bound optimizers include expectation-maximization (EM) algorithms, the concave-convex procedure (CCCP) \cite{Yuille2001} and submodular-supermodular procedures (SSP) \cite{Narasimhan2005}, among many others.}. At each iteration, it updates the variable as the minimum of a {\em surrogate function}, i.e., an upper bound on the original objective, which is tight at the current iteration. This guarantees that the original objective does not increase at each iteration.

Re-arranging the soft assignment matrix $\YY$ in vector form $\YY=[\yy_q] \in \mathbb{R}^{NC}$, relaxation $\tilde{\Lcal}(\YY)$ can be written conveniently in the following form:
\begin{equation}
\label{Laplacian-bound}
\tilde{\Lcal}(\YY) = - \sum_{q,p} w (\xxq_q, \xxq_p) \yy_q^t \yy_p = \YY^t\Psi \YY 
\end{equation}
with $\Psi = - \WW \otimes \II$, where $\otimes$ denotes the Kronecker product and $\II$ is the $N \times N$ identity matrix. Note that $\Psi$ is negative semi-definite for a positive semi-definite $\WW$. Therefore, $\YY^t\Psi \YY$ is a concave function, and the first-order approximation of \eqref{Laplacian-bound} at a current solution $\YY^i$ ($i$ is the iteration index) gives the following tight upper bound on $\tilde{\Lcal}(\YY)$:
\begin{equation}
\label{laplacian-upper-bound}
\tilde{\Lcal}(\YY) = \YY^t\Psi \YY  \leq (\YY^i)^t \Psi \YY^i + 2 \, (\Psi \YY^i)^t (\YY - \YY^i) 
\end{equation}

Therefore, using unary potentials $\Ncal(\YY)$ and the negative entropy barrier
in conjunction with the upper bound in \eqref{laplacian-upper-bound},
we obtain the following surrogate function $\Bcal_i(\YY)$ 
for relaxation $\Rcal(\YY)$ at current solution $\YY^i$:
\begin{equation}
\label{Aux-function}
\Rcal(\YY) \leq \Bcal_i(\YY)  \ceq  \sum_{q =1}^{N} \yy_q^t (\log (\yy_q) + {\mathbf a}_q - \lambda{\mathbf b}_q^i)
\end{equation}
where $\ceq$ means equality up to an additive constant\footnote{The additive constant in $\Bcal_i(\YY)$ is a term that depends only on $\YY^i$. This term comes from the Laplacian upper bound in \eqref{laplacian-upper-bound}.} that is independent of variable $\YY$, and 
${\mathbf a}_q$ and ${\mathbf b}_q^i$ are the following $C$-dimensional vectors:
\begin{subequations} 
\begin{align}
{\mathbf a}_q &=  [a_{q,1}, \dots,a_{q,C}]^t;  \, \, a_{q,c} = d(\xxq_q,\mm_c) \label{aq} \\
{\mathbf b}_q^i &= [b_{q,1}^i, \dots,b_{q,C}^i]^t;  \, \, b_{q,c}^i =   \sum_{p} w (\xxq_q, \xxq_p) y_{p,c}^i \label{bq}
\end{align}
\end{subequations}

It is straightforward to verify that upper bound $\Bcal_i(\YY)$ is tight at the current iteration, i.e., $\Bcal_i(\YY^i) = \Rcal(\YY^i)$.
This can be seen easily from the first-order approximation in \eqref{laplacian-upper-bound}. 
We iteratively optimize the surrogate function at each iteration $i$:
\begin{equation}
\label{bound_optimizer}
\YY^{i+1} = \argmin_{\YY} \Bcal_i(\YY)
\end{equation}
Because of upper-bound condition $\Rcal(\YY) \leq \Bcal_i(\YY), \forall \YY$, tightness condition $\Bcal_i(\YY^i) = \Rcal(\YY^i)$ at the current solution, and the fact that $\Bcal_i(\YY^{i+1}) \leq \Bcal_i(\YY^i)$ due to minimization \eqref{bound_optimizer}, it is easy to verify that updates \eqref{bound_optimizer} guarantee that relaxation ${\Rcal}(\YY)$ does not increase at each iteration: 
\[ \Rcal(\YY^{i+1})\leq \Bcal_i(\YY^{i+1}) \leq \Bcal_i(\YY^i) = \Rcal(\YY^i) \]

{\bf Closed-form solutions of the surrogate functions:} Notice that $\Bcal_i(\YY)$ is a sum of independent functions of each assignment variable. Therefore, we can 
solve \eqref{bound_optimizer} for each $\yy_q$ independently, while satisfying the simplex constraint:
\begin{equation}
\label{independant-Aux-function}
\min_{\yy_q \in \nabla_C} \yy_q^t (\log (\yy_q) + {\mathbf a}_q - \lambda{\mathbf b}_q^i), \, \forall q
\end{equation} 
The negative entropy barrier term $\yy_q^t \log \yy_q$ in \eqref{independant-Aux-function} restricts $\yy_q$ to be non-negative, removing the need of extra dual variables for the constraints $\yy_q>0$. Also, simplex constraint $\one^t\yy_q=1$ is affine. Thus, the solution of the following Karush-Kuhn-Tucker (KKT) condition provide the minimum of \eqref{independant-Aux-function}:
\begin{equation}
\label{KKT}
\log \yy_q + {\mathbf a}_q - \lambda{\mathbf b}_q^i + \beta \one = 0
\end{equation}
with $\beta$ the Lagrange multiplier for the simplex constraint. This provides, for each $q$, closed-form solutions for both the primal and dual variables, yielding the following independent updates of the assignment variables: 
\begin{equation}
\label{closed-form}
\yy_{q}^{i+1} = \frac{\exp ( - {\mathbf a}_q^i+ \lambda{\mathbf b}_q^i) }{{\mathbf 1}^t \exp (-{\mathbf a}_q^i+ \lambda{\mathbf b}_q^i)} \, \, \forall \, q 
\end{equation}

\begin{table*}[t]
\caption{Average accuracy (in \%) in \textit{mini}ImageNet and \textit{tiered}ImageNet. The best results are reported in bold font.}
\label{tab:miniimagenet}
\vskip 0.1in
\begin{center}
\begin{small}
\begin{tabular}{lc|cc|cc}
\toprule
& &\multicolumn{2}{c|}{\textbf{\textit{mini}ImageNet}}&\multicolumn{2}{c}{\textbf{\textit{tiered}ImageNet}} \\
\cline{3-6}
\textbf{Methods} & \textbf{Network}&\textbf{1-shot}& \textbf{5-shot}& \textbf{1-shot}& \textbf{5-shot}\\
\hline
MAML \cite{Finn2017ModelAgnosticMF} & ResNet-18 & 49.61 $\pm$ 0.92 & 65.72 $\pm$ 0.77&-&-\\
Chen \cite{chen2018a} & ResNet-18 & 51.87 $\pm$ 0.77 & 75.68 $\pm$ 0.63&-&-\\
RelationNet \cite{sung2018learning} & ResNet-18 & 52.48 $\pm$ 0.86 & 69.83 $\pm$ 0.68&-&-\\
MatchingNet \cite{Vinyals2016MatchingNF} & ResNet-18 & 52.91 $\pm$ 0.88 & 68.88 $\pm$ 0.69&-&-\\
ProtoNet \cite{snell2017prototypical} & ResNet-18 & 54.16 $\pm$ 0.82 & 73.68 $\pm$ 0.65&-&-\\
Gidaris \cite{gidaris2018dynamic}  & ResNet-15 & 55.45 $\pm$ 0.89 & 70.13 $\pm$ 0.68&-&-\\
SNAIL \cite{mishra2018a} & ResNet-15 & 55.71 $\pm$ 0.99 & 68.88 $\pm$ 0.92&-&-\\
AdaCNN \cite{pmlr-v80-munkhdalai18a} & ResNet-15 & 56.88 $\pm$ 0.62 & 71.94 $\pm$ 0.57&-&-\\
TADAM \cite{oreshkin2018tadam}  & ResNet-15 & 58.50 $\pm$ 0.30 & 76.70 $\pm$ 0.30&-&-\\
CAML \cite{jiang2018learning} & ResNet-12 & 59.23 $\pm$ 0.99 & 72.35 $\pm$ 0.71&-&-\\
TPN \cite{liu2018learning} & ResNet-12 & 59.46 & 75.64&-&-\\
TEAM \cite{team}& ResNet-18 & 60.07 & 75.90&-&-\\
MTL \cite{sun2019mtl}  & ResNet-18 & 61.20 $\pm$ 1.80 & 75.50 $\pm$ 0.80&-&-\\
VariationalFSL \cite{Variationalfewshot}  & ResNet-18 & 61.23 $\pm$ 0.26 & 77.69 $\pm$ 0.17&-&-\\
Transductive tuning \cite{Dhillon2020A} & ResNet-12 & 62.35 $\pm$ 0.66 & 74.53 $\pm$ 0.54&-&-\\
MetaoptNet \cite{lee2019meta}  & ResNet-18 & 62.64 $\pm$ 0.61 & 78.63 $\pm$ 0.46&65.99 $\pm$ 0.72 & 81.56 $\pm$ 0.53\\
SimpleShot \cite{wang2019simpleshot}& ResNet-18 & 63.10 $\pm$ 0.20 & 79.92 $\pm$ 0.14& 69.68 $\pm$ 0.22 & 84.56 $\pm$ 0.16\\
CAN+T \cite{can}& ResNet-12 & 67.19 $\pm$ 0.55 & 80.64 $\pm$ 0.35& 73.21 $\pm$ 0.58 & 84.93 $\pm$ 0.38\\
\rowcolor{Gray}
LaplacianShot (ours) & ResNet-18 & \textbf{72.11} $\pm$ 0.19 & \textbf{82.31} $\pm$ 0.14& \textbf{78.98} $\pm$ 0.21 & \textbf{86.39} $\pm$ 0.16\\
\hline
Qiao \cite{qiao2018few}  & WRN & 59.60 $\pm$ 0.41 & 73.74 $\pm$ 0.19&-&-\\
LEO \cite{rusu2018metalearning} & WRN & 61.76 $\pm$ 0.08 &77.59 $\pm$ 0.12& 66.33 $\pm$ 0.05 & 81.44 $\pm$ 0.09\\
ProtoNet \cite{snell2017prototypical} & WRN & 62.60 $\pm$ 0.20 & 79.97 $\pm$ 0.14&-&-\\
CC+rot \cite{gidaris2019boosting}  & WRN & 62.93 $\pm$ 0.45 & 79.87 $\pm$ 0.33& 70.53 $\pm$ 0.51 & 84.98 $\pm$ 0.36\\
MatchingNet \cite{Vinyals2016MatchingNF} & WRN & 64.03 $\pm$ 0.20 & 76.32 $\pm$ 0.16&-&-\\
FEAT \cite{ye2020fewshot} & WRN & 65.10 $\pm$ 0.20 & 81.11 $\pm$ 0.14& 70.41 $\pm$ 0.23 & 84.38 $\pm$ 0.16\\
Transductive tuning \cite{Dhillon2020A} & WRN & 65.73 $\pm$ 0.68 & 78.40 $\pm$ 0.52& 73.34 $\pm$ 0.71 & 85.50 $\pm$ 0.50\\
SimpleShot \cite{wang2019simpleshot}& WRN & 65.87$\pm$ 0.20 & 82.09 $\pm$ 0.14& 70.90 $\pm$ 0.22 & 85.76 $\pm$ 0.15\\
SIB  \cite{hu2020empirical} & WRN & 70.0 $\pm$ 0.6 & 79.2 $\pm$ 0.4&-&-\\
BD-CSPN \cite{liu2019prototype} & WRN & 70.31 $\pm$ 0.93 & 81.89 $\pm$ 0.60& 78.74 $\pm$ 0.95 & 86.92 $\pm$ 0.63\\

\rowcolor{Gray}
LaplacianShot (ours) & WRN & \textbf{74.86} $\pm$ 0.19 & \textbf{84.13} $\pm$ 0.14& \textbf{80.18} $\pm$ 0.21 & \textbf{87.56}$\pm$ 0.15\\
\hline
SimpleShot \cite{wang2019simpleshot}& MobileNet & 61.55 $\pm$ 0.20 & 77.70 $\pm$ 0.15& 69.50 $\pm$ 0.22 & 84.91 $\pm$ 0.15\\
\rowcolor{Gray}
LaplacianShot (ours) & MobileNet & \textbf{70.27} $\pm$ 0.19 & \textbf{80.10} $\pm$ 0.15& \textbf{79.13} $\pm$ 0.21 & \textbf{86.75} $\pm$ 0.15\\
\hline
SimpleShot \cite{wang2019simpleshot}& DenseNet& 65.77 $\pm$ 0.19 & 82.23 $\pm$ 0.13&71.20 $\pm$ 0.22 & 86.33 $\pm$ 0.15\\
\rowcolor{Gray}
LaplacianShot (ours) & DenseNet & \textbf{75.57} $\pm$ 0.19 & \textbf{84.72} $\pm$ 0.13& \textbf{80.30} $\pm$ 0.22 & \textbf{87.93} $\pm$ 0.15\\
\bottomrule
\end{tabular}
\end{small}
\end{center}
\vskip -0.1in
\end{table*}

\begin{table*}[t]
\caption{Results for CUB and cross-domain results on \textbf{\textit{mini}Imagenet} $\rightarrow$ \textbf{CUB}.}
\label{tab:cub}
\vskip 0.1in
\begin{center}
\begin{small}
\begin{tabular}{lccccr}
\toprule
    \multirow{2}{*}{\textbf{Methods}}& \multirow{2}{*}{\textbf{Network}}& \multicolumn{2}{c}{\textbf{CUB}} & \multicolumn{2}{c}{\textbf{\textit{mini}Imagenet} $\rightarrow$ \textbf{CUB}}\\
     & & \textbf{1-shot}& \textbf{5-shot} & \textbf{1-shot}&\textbf{5-shot}\\
     \toprule
     MatchingNet \cite{Vinyals2016MatchingNF}& ResNet-18 &  73.49 & 84.45 & - & 53.07\\
     MAML \cite{Finn2017ModelAgnosticMF} & ResNet-18 & 68.42 & 83.47  & - & 51.34 \\
     ProtoNet \cite{snell2017prototypical} & ResNet-18 & 72.99 & 86.64  & - & 62.02 \\
     RelationNet \cite{sung2018learning} & ResNet-18 & 68.58 & 84.05  & - & 57.71 \\
     Chen \cite{chen2018a} & ResNet-18 & 67.02 & 83.58  & - & 65.57 \\
     SimpleShot \cite{wang2019simpleshot} & ResNet-18 & 70.28 & 86.37 & 48.56 &65.63\\
     \midrule
     LaplacianShot(ours) & ResNet-18 & \textbf{80.96} & \textbf{88.68} &\textbf{55.46} &\textbf{66.33}\\
     \bottomrule
\end{tabular}
\end{small}
\end{center}
\vskip -0.1in
\end{table*}
\begin{table*}[tb]
\caption{Average accuracy (in \%) in iNat benchmark for SimpleShot \cite{wang2019simpleshot} and the proposed LaplacianShot. The best results are reported in bold font. Note that, for iNat, we do not utilize the rectified prototypes. [The best reported result of \cite{wertheimer2019few} with ResNet50 is: Per Class: 46.04\%, Mean: 51.25\%.]}
\label{tab:iNat}
\vskip 0.1in
\begin{center}
\begin{small}
\begin{tabular}{lccccccr}
\toprule
\multirow{2}{*}{\textbf{Methods}} & \multirow{2}{*}{\textbf{Network}} & \multicolumn{2}{c}{\textbf{UN}}& \multicolumn{2}{c}{\textbf{L2}}&\multicolumn{2}{c}{\textbf{CL2}}\\
&&\textbf{Per Class}&\textbf{Mean}&\textbf{Per Class}&\textbf{Mean}&\textbf{Per Class}&\textbf{Mean}\\
\midrule
SimpleShot & ResNet-18 &55.80&58.56&57.15& 59.56 & 56.35& 58.63\\
\rowcolor{Gray}
LaplacianShot & ResNet-18 &\textbf{62.80}&\textbf{66.40}&\textbf{58.72}&\textbf{61.14}&\textbf{58.49}&\textbf{60.81}\\
\hline
SimpleShot & ResNet-50 &58.45&61.07&59.68&61.99&58.83&60.98\\
\rowcolor{Gray}
LaplacianShot & ResNet-50 &\textbf{65.96}&\textbf{69.13}&\textbf{61.40}&\textbf{63.66}&\textbf{61.08}&\textbf{63.18}\\
\hline
SimpleShot & WRN &62.44&65.08&64.26&66.25&63.03&65.17\\
\rowcolor{Gray}
LaplacianShot &WRN &\textbf{71.55} & \textbf{74.97}& \textbf{65.78}& \textbf{67.82}&\textbf{65.32} &\textbf{67.43}\\
\bottomrule
\end{tabular}
\end{small}
\end{center}
\vskip -0.1in
\end{table*}
\subsection{Proposed Algorithm}
The overall proposed algorithm is simplified in Algorithm \ref{alg}. Once the network function $\ftheta$ is learned using the base dataset $\XXb$, our algorithm proceeds with the extracted features $\xxq_q$. Before the iterative bound updates, each 
soft assignment $\yy_q^1$ is initialized as a softmax probability of $\mathbf{a}_q$, which is based on the distances to prototypes $\mm_c$. The iterative bound optimization is guaranteed to converge, typically less than $15$ iterations in our experiments (Figure \ref{fig:convergence}). Also the independent point-wise bound updates yield a parallel structure of the algorithm, which makes it very efficient (and convenient for large-scale few-shot tasks). We refer to our method as 
LaplacianShot in the experiments.

\textbf{Link to attention mechanisms:} Our Laplacian-regularized model has interesting connection to the popular attention mechanism in \cite{Vaswani17}. In fact, MatchingNet \cite{Vinyals2016MatchingNF} predicted the labels of the query samples $\xxq_q$ as a linear combination of the support labels. The expression of $b_{q,c}^i$ that we obtained in \eqref{bq}, which stems from our bound optimizer and the concave relaxation of the Laplacian, also takes the form of a combination of labels at each iteration $i$ in our model: $b_{q,c}^i =   \sum_{p} w (\xxq_q, \xxq_p) y_{p,c}^i$. However, there are
important differences with \cite{Vinyals2016MatchingNF}: First, the attention in our formulation is non-parametric as it considers only the feature relationships among the query samples in $\XXq$, not the support examples. Second, unlike our approach, the attention mechanism in \cite{Vinyals2016MatchingNF} is employed during training for learning embedding function $\ftheta$ with a meta-learning approach.
\begin{figure*}
\vskip 0.1in
\begin{center}
\includegraphics[width=\textwidth]{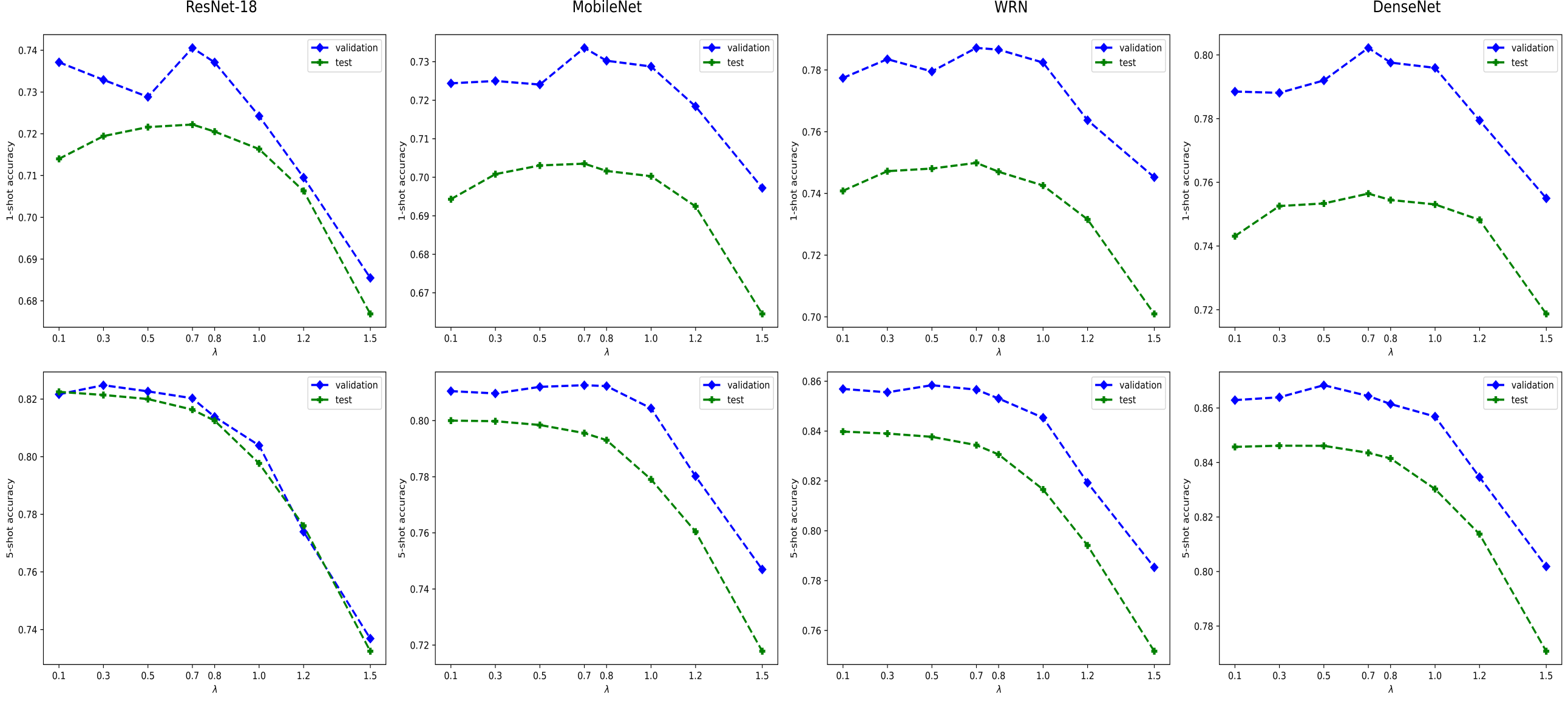}
\caption{ We tune regularization parameter $\lambda$ over values ranging from 0.1 to 1.5. In the above plots, we show the impact of choosing $\lambda$ on both validation and test accuracies. The values of $\lambda$ based on the best validation accuracies correspond to good accuracies in the test classes. The results are shown for different networks on \textit{mini}ImageNet dataset, for both 1-shot (top row) and 5-shot (bottom row).}
\label{fig:lambda}
\end{center}
\vskip -0.2in
\end{figure*}
\section{Experiments}
In this section, we describe our experimental setup. An implementation of our LaplacianShot is publicly available\footnote{\hyperlink{https://github.com/imtiazziko/LaplacianShot}{https://github.com/imtiazziko/LaplacianShot}}.
\subsection{Datasets}
We used five benchmarks for few-shot classification: 
\textit{mini}ImageNet, \textit{tiered}ImageNet, CUB, cross-domain CUB (with base training on \textit{mini}ImageNet) and iNat.

The \textbf{\textit{mini}ImageNet} benchmark is a subset of the larger ILSVRC-12 dataset \cite{ILSVRC15}. It has a total of 60,000 color images with 100 classes, where each class has 600 images of size $84 \times 84$, following \cite{Vinyals2016MatchingNF}. We use the standard split of 64 base, 16 validation and 20 test classes \cite{Ravi2017OptimizationAA,wang2019simpleshot}. The \textbf{\textit{tiered}ImageNet} benchmark \cite{ren18fewshotssl} is also a subset of ILSVRC-12 dataset but with 608 classes instead. We follow standard splits with 351 base, 97 validation and 160 test classes for the experiments. The images are also resized to $84 \times 84$ pixels. \textbf{CUB}-200-2011 \cite{wah2011caltech} is a fine-grained image classification dataset. We follow \cite{chen2018a} for few-shot classification on CUB, which splits into 100 base, 50 validation and 50 test classes for the experiments. The images are also resized to $84 \times 84$ pixels, as in \textit{mini}ImageNet. The \textbf{iNat} benchmark, introduced recently for few-shot classification in \cite{wertheimer2019few}, contains images of 1,135 animal species. It introduces a more challenging few-shot scenario, with different numbers of support examples per class, which simulates more realistic class-imbalance scenarios, and with semantically related classes that are not easily separable. Following \cite{wertheimer2019few}, the dataset is split into 908 base classes and 227 test classes, with images of size $84 \times 84$.

\subsection{Evaluation Protocol}
In the case of \textbf{\textit{mini}ImageNet}, \textbf{CUB} and \textbf{\textit{tiered}ImageNet}, we evaluate 10,000 five-way 1-shot and five-way 5-shot classification tasks, randomly sampled from the test classes, following standard few-shot evaluation settings \cite{wang2019simpleshot,rusu2018metalearning}. This means that, for each of the five-way few-shot tasks, $C=5$ classes are randomly selected, with $|\XXs^c|=1$ (1-shot) and $|\XXs^c|=5$ (5-shot) examples selected per class, to serve as support set $\XXs$.
Query set $\XXq$ contains 15 images per class. Therefore, the evaluation is performed over $N=75$ query images per task. The average accuracy of these 10,000 few shot tasks are reported along with the 95\% confidence interval. For the \textbf{iNat} benchmark, the number of support examples $|\XXs^c|$ per class varies. We performed 227-way multi-shot evaluation, and report the top-1 accuracy averaged over the test images per class (Per Class in Table \ref{tab:iNat}), as well as the average over all test images (Mean in Table \ref{tab:iNat}), following the same procedure as in \cite{wertheimer2019few,wang2019simpleshot}. 

\subsection{Network Models}
We evaluate LaplacianShot on four different backbone network models to learn feature extractor $\ftheta$: 

\textbf{ResNet-18/50} is based on the deep residual network architecture \cite{he2016deep}, where the first two down-sampling layers are removed, setting the stride to 1 in the first convolutional layer and removing the first max-pool layer. The first convolutional layer is used with a kernel of size $3 \times 3$ instead of $7 \times 7$. ResNet-18 has 8 basic residual blocks, and  ResNet-50 has 16 bottleneck blocks. For all the networks, the dimension of the extracted features is 512. \textbf{MobileNet} \cite{mobilenet} was initially proposed as a light-weight convolutional network for mobile-vision applications. In our setting, we remove the first two down-sampling operations, which results in a feature embedding of size 1024. \textbf{WRN} \cite{WRN} widens the residual blocks by adding more convolutional layers and feature planes. In our case, we used 28 convolutional layers, with a widening factor of 10 and an extracted-feature dimension of 640. Finally, we used the standard 121-layer \textbf{DenseNet} \cite{densenet}, omitting the first two down-sampling layers and setting the stride to 1. We changed the kernel size of the first convolutional layer to $3 \times 3$. The extracted feature vector is of dimension 1024.

\begin{figure*}[tb]
\vskip 0.1in
\begin{center}
\centerline{\includegraphics[width=\textwidth]{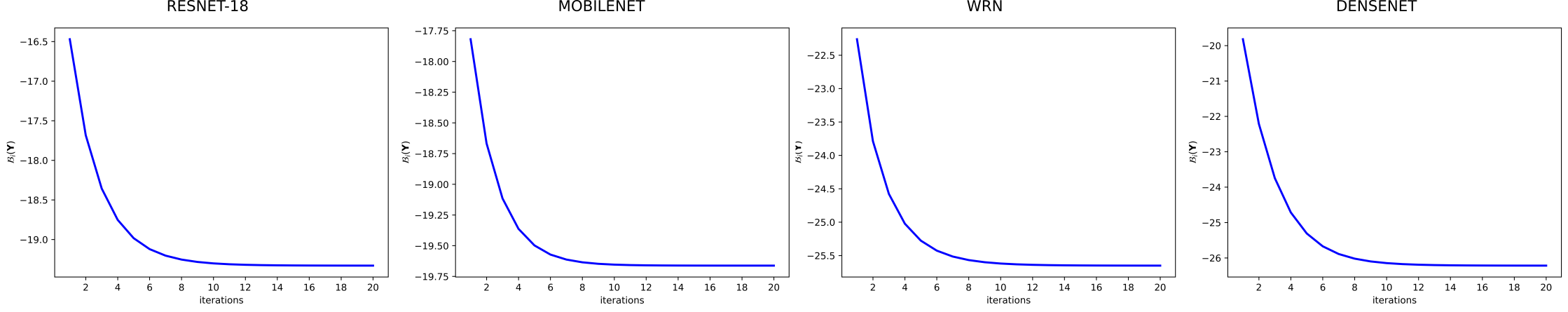}}
\caption{Convergence of Algorithm \ref{alg}:  Bounds $\Bcal_i(\YY)$ vs. iteration numbers for features from different networks. Here, the plots are produced by setting $\lambda=1.0$, for a single $5$-way $5$ shot task from the \textit{mini}ImageNet test set.}
\label{fig:convergence}
\end{center}
\vskip -0.1in
\end{figure*}

\subsection{Implementation Details}
\textbf{Network model training:} We trained the network models using the standard cross-entropy loss on the base classes, with a label-smoothing \cite{szegedy2016rethinking} parameter set to $0.1$.
Note that the base training did not involve any meta-learning or episodic-training strategy. 
We used the SGD optimizer to train the models, with  
mini-batch size set to 256 for all the networks, except for WRN and DenseNet, where we used mini-batch sizes of 128 and 100, respectively. We used two 16GB P100 GPUs for network training with base classes. 
For \textit{mini}ImageNet, CUB and \textit{tiered}ImageNet, we used early stopping by evaluating the the nearest-prototype classification accuracy on the validation classes, with \textbf{L2} normalized features.

\textbf{Prototype estimation and feature transformation:}
During the inference on test classes, SimpleShot \cite{wang2019simpleshot} performs the following feature transformations: \textbf{L2} normalization, $\xxq_q:=\xxq_q/\|\xxq_q\|_2$ and \textbf{CL2}, which computes the mean of the base class features $\bar{\xx} = \frac{1}{|\XXb|}\sum_{\xx \in\XXb} \xx$ and centers the extracted features as $\xxq_q:= \xxq_q -\bar{\xx}$, which is followed by an L2 normalization. We report the results in Table \ref{tab:miniimagenet} and \ref{tab:cub} with CL2 normalized features. In Table \ref{tab:iNat} for the iNat dataset, we provide the results with both normalized and unnormalized (\textbf{UN}) features for a comparative analysis. We reproduced the results of SimpleShot with our trained network models. In the \textbf{1-shot} setting, prototype $\mm_c$ is just the support example $\xxq_q \in \XXs^c$ of class c, whereas in \textbf{multi-shot}, $\mm_c$ is the simple mean of the support examples of class $c$. Another option is to use \emph{rectified} prototypes, i.e., a weighted combination of features from both the support examples in $\XXs^c$ and query samples in $\XXq^c$, which are initially predicted as belonging to class $c$ using Eq. \eqref{eq:initial_pred}: 
\[\tilde{\mm}_c = \frac{1}{|\XXs^c|+|\XXq^c|}\sum_{\xxq_p \in \{\XXs^c,\XXq^c\}}  \frac{\exp(cos(\xxq_p,\mm_c))}{\sum_{c=1}^{C} \exp(cos(\xxq_p,\mm_c))}\xxq_p,\]
where $cos$ denotes the cosine similarity. And, for a given few-shot task, we compute the cross-domain shift $\Delta$ as the difference between the mean of features within the support set and the mean of features within the query set: $\Delta = \frac{1}{|\XXs|}\sum_{\xx_p \in\XXs} \xx_p - \frac{1}{|\XXq|}\sum_{\xx_q \in\XXq} \xx_q$. 
Then, we rectify each query point $\xxq_p \in \XXq$ in the few-shot task as follows: $\xxq_p = \xxq_p + \Delta$. This shift correction is similar to the prototype rectification in \cite{liu2019prototype}.
Note that our LaplacianShot model in Eq. \eqref{eq:LN} is agnostic to the way of estimating the prototypes: It can be used either with the standard prototypes ($\mm_c$) or with the rectified ones ($\tilde{\mm}_{c}$). 
We report the results of LaplacianShot with the rectified prototypes in Table \ref{tab:miniimagenet} and \ref{tab:cub}, for \textit{mini}Imagenet, \textit{tiered}Imagenet and CUB. We do not report the results with the rectified prototypes in Table \ref{tab:iNat} for iNat, as rectification drastically worsen the performance.

For $\WW$, we used the k-nearest neighbor affinities as follows: $w(\xxq_q,\xxq_p) =1$ if $\xxq_p$ is within the $k$ nearest neighbor of $\xxq_q$, and $w(\xxq_q,\xxq_p) =0$ otherwise. In our experiments, $k$ is simply chosen from three typical values (3, 5 or 10) tuned over 500 few-shot tasks from the base training classes (i.e., we did not use test data for choosing $k$). We used $k=3$ for \textit{mini}ImageNet, CUB and \textit{tiered}ImageNet and $k=10$ for iNat benchmark. Regularization parameter $\lambda$ is chosen based on the validation class accuracy for \textit{mini}ImageNet, CUB and \textit{tiered}ImageNet. This will be discussed in more details in section \ref{lambda}. For the iNat experiments, we simply fix $\lambda=1.0$, as there is no validation set for this benchmark.

\begin{table*}[t]
\caption{Ablation study on the effect of each term corresponding to nearest prototype $\Ncal(\YY)$, Laplacian $\Lcal(\YY)$ and rectified prototype $\tilde{\mm}_{c}$. Results are reported with ResNet-18 network. Note that, the Laplacian regularization $\Lcal(\YY)$ improve the results consistently.}
\label{tab:ablation}
\vskip 0.1in
\begin{center}
\begin{small}
       \begin{tabular}{ccccccccc}
             & & & \multicolumn{2}{c}{\textbf{\textit{mini}-ImageNet}} & \multicolumn{2}{c}{\textbf{\textit{tiered}-ImageNet}} & \multicolumn{2}{c}{\textbf{CUB}}\\ $\Ncal(\YY)$ & $\Lcal(\YY)$ &$\tilde{\mm}_{c}$ & 1-shot & 5-shot & 1-shot & 5-shot & 1shot & 5-shot \\
             \toprule
             \cmark & \xmark & \xmark & 63.10 & 79.92 & 69.68 & 84.56 & 70.28 & 86.37 \\
             \rowcolor{Gray}
             \cmark & \cmark & \xmark & 66.20 & 80.75 & 72.89 & 85.25 & 74.46 & 86.86 \\
             \cmark & \xmark & \cmark & 69.74 & 82.01 & 76.73 & 85.74 & 78.76 & 88.55 \\
             \rowcolor{Gray}
             \cmark & \cmark & \cmark & \textbf{72.11} & \textbf{82.31} & \textbf{78.98} & \textbf{86.39} & \textbf{80.96} & \textbf{88.68} \\
             \bottomrule
        \end{tabular}
    \end{small}
    \end{center}
    \vskip -0.1in
    \end{table*}
\subsection{Results}
\label{results}
We evaluated LaplacianShot over five different benchmarks, with different scenarios and difficulties:
Generic image classification, fine-grained image classification, cross-domain adaptation, and imbalanced class distributions.
We report the results of LaplacianShot for \textit{mini}ImageNet, \textit{tiered}ImageNet, CUB and iNat datasets, in Tables \ref{tab:miniimagenet}, \ref{tab:cub} and \ref{tab:iNat}, along with 
comparisons with state-of-the-art methods.  

\textbf{Generic image classification}: Table \ref{tab:miniimagenet} reports the results of generic image classification for the standard \textit{mini}ImageNet and \textit{tiered}ImageNet few-shot benchmarks. We can clearly observe that LaplacianShot outperforms state-of-the-art methods by large margins, with gains that are consistent across different settings and network models. It is worth mentioning that, for challenging scenarios, e.g., 1-shot with low-capacity models, LaplacianShot outperforms complex meta-learning methods by more than 9\%. For instance, compared to well-known MAML \cite{Finn2017ModelAgnosticMF} and ProtoNet \cite{snell2017prototypical}, and to the recent MetaoptNet \cite{lee2019meta}, LaplacianShot brings improvements of nearly 22\%, 17\%, and 9\%, respectively, under the same evaluation conditions. Furthermore, it outperforms the very recent transductive approaches in \cite{Dhillon2020A,liu2019prototype,liu2018learning} by significant margins. 
With better learned features with WRN and DenseNet, LaplacianShot brings significant performance boosts, yielding state-of-the art results in few-shot classification, without meta-learning. 

\textbf{Fine-grained image classification}:  Table \ref{tab:cub} reports the results of fine-grained few-shot classification on CUB, with Resnet-18 network. LaplacianShot outperforms the best performing method in this setting by a 7\% margin.

\textbf{Cross-domain (mini-ImageNet $\rightarrow$ CUB)}: We perform the very interesting few-shot experiment, with a cross-domain scenario, following the setting in \cite{chen2018a}. We used the ResNet-18 model trained on the \textit{mini}Imagenet base classes, while evaluation is performed on CUB few-shot tasks, with 50 test classes. Table \ref{tab:cub} (rightmost column) reports the results. In this cross-domain setting, and consistently with the standard settings, LaplacianShot  outperforms complex meta-learning methods by substantial margins. 

\textbf{Imbalanced class distribution:} Table \ref{tab:iNat} reports the results for the more challenging, class-imbalanced iNat benchmark, with different numbers of support examples per class and, also, with high visual similarities between the different classes, making class separation difficult. To our knowledge, only \cite{wertheimer2019few,wang2019simpleshot} report performances on this benchmark, and SimpleShot \cite{wang2019simpleshot} represents the state-of-the-art. We compared with SimpleShot using unnormalized extracted features (UN), L2 and CL2 normalized features. Our Laplacian regularization yields significant improvements, regardless of the network model and feature normalization. However, unlike SimpleShot, our method reaches its best performance with the unnormalized features. Note that, for iNat, we did not use the rectified prototypes. These results clearly highlight the benefit Laplacian regularization brings in challenging class-imbalance scenarios.
\begin{table}
\caption{Average inference time (in seconds) for the 5-shot tasks in \textit{mini}Imagenet dataset.}
\label{tab:timing}
\vskip 0.1in
\begin{center}
\begin{small}
\resizebox{\linewidth}{!}{
\begin{tabular}{lc}
\toprule
\textbf{Methods}& \textbf{inference time} \\
\midrule
SimpleShot \cite{wang2019simpleshot}&0.009 \\
Transductive tuning \cite{Dhillon2020A}&20.7 \\
\rowcolor{Gray}
LaplacianShot (ours) &  0.012  \\
\bottomrule
\end{tabular}
}
\end{small}
\end{center}
\vskip -0.25in
\end{table}
\subsection{Ablation Study}
\label{lambda}
\textbf{Choosing the Value of $\lambda$:}
In LaplacianShot, we need to choose the value of regularization parameter $\lambda$, which controls the trade-off between the nearest-prototype classifier term ${\mathbf a}_q$ and Laplacian regularizer ${\mathbf b}_q^i$. We tuned this parameter using the validation classes by sampling $500$ few-shot tasks. LaplacianShot is used in each few-shot task with the following values of $\lambda$: $[0.1,0.3,0.5,0.7,0.8,1.0,1.2,1.5]$. The best $\lambda$ corresponding to the best average 1-shot and 5-shot accuracy over validation classes/data is selected for inference over the test classes/data. To examine experimentally whether the chosen values of $\lambda$ based on the best validation accuracies correspond to good accuracies in the test classes, we plotted both the validation and test class accuracies 
vs. different values of $\lambda$ for \textit{mini}ImageNet (Figure \ref{fig:lambda}). The results are intuitive, with a consistent trend in both 1-shot and 5-shot settings. Particularly, for 1-shot tasks, $\lambda=0.7$ provides the best results in both  validation and test accuracies. In 5-shot tasks, the best test results are obtained mostly with $\lambda=0.1$, while the best validation accuracies were reached with higher values of $\lambda$. Nevertheless, we report the results of LaplacianShot with the values of $\lambda$ chosen based on the best validation accuracies.

\textbf{Effects of Laplacian regularization:}
We conducted an ablation study on the effect of each term in our model, i.e., nearest-prototype classifier $\Ncal(\YY)$ and Laplacian regularizer $\Lcal(\YY)$. We also examined the effect of using prototype rectification, i.e., $\tilde{\mm}_{c}$ instead of $\mm_{c}$.
Table \ref{tab:ablation} reports the results, using the ResNet-18 network. The first row corresponds to the prediction of the nearest neighbor classifier ($\lambda=0$), and the second shows the effect of adding Laplacian regularization. In the 1-shot case, the latter boosts the performances by at least 3\%. Prototype rectification (third and fourth rows) also boosts the performances. Again, in this case, the improvement that the Laplacian term brings is significant, particularly in the 1-shot case (2 to 3\%).

\textbf{Convergence of transductive LaplacianShot inference}:
The proposed algorithm belongs to the family of bound optimizers or MM algorithms. 
In fact, the MM principle can be viewed as a generalization of expectation-maximization (EM). Therefore, in general, MM algorithms inherit the monotonicity and convergence properties of EM algorithms \cite{vaida2005parameter}, which are well-studied in the literature. In fact, Theorem 3 in \cite{vaida2005parameter} states a simple condition for convergence of the general MM procedure, which is almost always satisfied in practice: The surrogate function has a unique global minimum. In Fig. \ref{fig:convergence}, we plotted surrogates $\Bcal_i(\YY)$, up to a constant, i.e., Eq. \eqref{Aux-function}, as functions of the iteration numbers, for different networks. One can see that the value of $\Bcal_i(\YY)$ decreases monotonically at each iteration, and converges, typically, within less than 15 iterations.

\textbf{Inference time}: We computed the average inference time required for each 5-shot task. Table \ref{tab:timing} reports these inference times for \textit{mini}ImageNet with the WRN network. The purpose of this is to check whether there exist a significant computational overhead added by our Laplacian-regularized transductive inference, in comparison to inductive inference. Note that the computational complexity of the proposed inference is $\mathcal{O}(NkC)$ for a few-shot task, where $k$ is the neighborhood size for affinity matrix $\WW$. The inference time per few-shot task for LaplacianShot is close to inductive SimpleShot run-time (LaplacianShot is only 1-order of magnitude slower), and is 3-order-of-magnitude faster than the transductive fine-tuning in \cite{Dhillon2020A}. 

\section{Conclusion}
Without meta-learning, we provide state-of-the-art results, outperforming significantly a large number of sophisticated few-shot learning methods, in all benchmarks. Our transductive inference is a simple constrained graph clustering of the query features. It can be used in conjunction with any base-class training model, consistently yielding improvements. Our results are in line with several recent baselines \cite{Dhillon2020A,chen2018a,wang2019simpleshot} that reported competitive performances, without resorting to complex meta-learning strategies. This recent line of simple methods emphasizes the limitations of current few-shot benchmarks, and questions the viability of a large body of convoluted few-shot learning techniques in the recent literature. As pointed out in Fig. 1 in \cite{Dhillon2020A}, the progress made by an abundant recent few-shot literature, mostly based on meta-learning, may be illusory. Classical and simple regularizers, such as the entropy in \cite{Dhillon2020A} or our Laplacian term, well-established in semi-supervised learning and clustering, achieve outstanding performances. We do not claim to hold the ultimate solution for few-shot learning, but we believe that our model-agnostic transductive inference should be used as a strong baseline for future few-shot learning research.

\bibliographystyle{icml2020}
\bibliography{readings.bib}

\begin{thebibliography}{52}
\providecommand{\natexlab}[1]{#1}
\providecommand{\url}[1]{\texttt{#1}}
\expandafter\ifx\csname urlstyle\endcsname\relax
  \providecommand{\doi}[1]{doi: #1}\else
  \providecommand{\doi}{doi: \begingroup \urlstyle{rm}\Url}\fi

\bibitem[Belkin et~al.(2006)Belkin, Niyogi, and Sindhwani]{belkin2006manifold}
Belkin, M., Niyogi, P., and Sindhwani, V.
\newblock Manifold regularization: A geometric framework for learning from
  labeled and unlabeled examples.
\newblock \emph{Journal of Machine Learning Research}, 7:\penalty0 2399--2434,
  2006.

\bibitem[Chen et~al.(2019)Chen, Liu, Kira, Wang, and Huang]{chen2018a}
Chen, W.-Y., Liu, Y.-C., Kira, Z., Wang, Y.-C.~F., and Huang, J.-B.
\newblock A closer look at few-shot classification.
\newblock In \emph{International Conference on Learning Representations
  (ICLR)}, 2019.

\bibitem[Dhillon et~al.(2020)Dhillon, Chaudhari, Ravichandran, and
  Soatto]{Dhillon2020A}
Dhillon, G.~S., Chaudhari, P., Ravichandran, A., and Soatto, S.
\newblock A baseline for few-shot image classification.
\newblock In \emph{International Conference on Learning Representations
  (ICLR)}, 2020.

\bibitem[Fei-Fei et~al.(2006)Fei-Fei, Fergus, and Perona]{fei2006one}
Fei-Fei, L., Fergus, R., and Perona, P.
\newblock One-shot learning of object categories.
\newblock \emph{IEEE Transactions on Pattern Analysis and Machine
  Intelligence}, 28:\penalty0 594--611, 2006.

\bibitem[Finn et~al.(2017)Finn, Abbeel, and Levine]{Finn2017ModelAgnosticMF}
Finn, C., Abbeel, P., and Levine, S.
\newblock Model-agnostic meta-learning for fast adaptation of deep networks.
\newblock In \emph{International Conference on Machine Learning (ICML)}, 2017.

\bibitem[Gidaris \& Komodakis(2018)Gidaris and Komodakis]{gidaris2018dynamic}
Gidaris, S. and Komodakis, N.
\newblock Dynamic few-shot visual learning without forgetting.
\newblock In \emph{Conference on Computer Vision and Pattern Recognition
  (CVPR)}, 2018.

\bibitem[Gidaris et~al.(2019)Gidaris, Bursuc, Komodakis, P{\'e}rez, and
  Cord]{gidaris2019boosting}
Gidaris, S., Bursuc, A., Komodakis, N., P{\'e}rez, P., and Cord, M.
\newblock Boosting few-shot visual learning with self-supervision.
\newblock In \emph{International Conference on Computer Vision (ICCV)}, 2019.

\bibitem[He et~al.(2016)He, Zhang, Ren, and Sun]{he2016deep}
He, K., Zhang, X., Ren, S., and Sun, J.
\newblock Deep residual learning for image recognition.
\newblock In \emph{Conference on Computer Vision and Pattern Recognition
  (CVPR)}, 2016.

\bibitem[Hou et~al.(2019)Hou, Chang, Bingpeng, Shan, and Chen]{can}
Hou, R., Chang, H., Bingpeng, M., Shan, S., and Chen, X.
\newblock Cross attention network for few-shot classification.
\newblock In \emph{Neural Information Processing Systems (NeurIPS)}, 2019.

\bibitem[Howard et~al.(2017)Howard, Zhu, Chen, Kalenichenko, Wang, Weyand,
  Andreetto, and Adam]{mobilenet}
Howard, A.~G., Zhu, M., Chen, B., Kalenichenko, D., Wang, W., Weyand, T.,
  Andreetto, M., and Adam, H.
\newblock Mobilenets: Efficient convolutional neural networks for mobile vision
  applications.
\newblock \emph{Preprint arXiv:1704.04861}, 2017.

\bibitem[Hu et~al.(2020)Hu, Moreno, Xiao, Shen, Obozinski, Lawrence, and
  Damianou]{hu2020empirical}
Hu, S.~X., Moreno, P.~G., Xiao, Y., Shen, X., Obozinski, G., Lawrence, N.~D.,
  and Damianou, A.
\newblock Empirical bayes transductive meta-learning with synthetic gradients.
\newblock In \emph{International Conference on Learning Representations
  (ICLR)}, 2020.

\bibitem[Huang et~al.(2017)Huang, Liu, Van Der~Maaten, and
  Weinberger]{densenet}
Huang, G., Liu, Z., Van Der~Maaten, L., and Weinberger, K.~Q.
\newblock Densely connected convolutional networks.
\newblock In \emph{Conference on Computer Vision and Pattern Recognition
  (CVPR)}, 2017.

\bibitem[Jiang et~al.(2019)Jiang, Havaei, Varno, Chartrand, Chapados, and
  Matwin]{jiang2018learning}
Jiang, X., Havaei, M., Varno, F., Chartrand, G., Chapados, N., and Matwin, S.
\newblock Learning to learn with conditional class dependencies.
\newblock In \emph{International Conference on Learning Representations
  (ICLR)}, 2019.

\bibitem[Kim et~al.(2019)Kim, Kim, Kim, and Yoo]{kim2019edge}
Kim, J., Kim, T., Kim, S., and Yoo, C.~D.
\newblock Edge-labeling graph neural network for few-shot learning.
\newblock In \emph{Conference on Computer Vision and Pattern Recognition
  (CVPR)}, 2019.

\bibitem[Lange et~al.(2000)Lange, Hunter, and Yang]{lange2000optimization}
Lange, K., Hunter, D.~R., and Yang, I.
\newblock Optimization transfer using surrogate objective functions.
\newblock \emph{Journal of computational and graphical statistics}, 9\penalty0
  (1):\penalty0 1--20, 2000.

\bibitem[Lee et~al.(2019)Lee, Maji, Ravichandran, and Soatto]{lee2019meta}
Lee, K., Maji, S., Ravichandran, A., and Soatto, S.
\newblock Meta-learning with differentiable convex optimization.
\newblock In \emph{Conference on Computer Vision and Pattern Recognition
  (CVPR)}, 2019.

\bibitem[Liu et~al.(2019{\natexlab{a}})Liu, Song, and Qin]{liu2019prototype}
Liu, J., Song, L., and Qin, Y.
\newblock Prototype rectification for few-shot learning.
\newblock \emph{Preprint arXiv:1911.10713}, 2019{\natexlab{a}}.

\bibitem[Liu et~al.(2019{\natexlab{b}})Liu, Lee, Park, Kim, Yang, Hwang, and
  Yang]{liu2018learning}
Liu, Y., Lee, J., Park, M., Kim, S., Yang, E., Hwang, S., and Yang, Y.
\newblock Learning to propagate labels: Transductive propagation network for
  few-shot learning.
\newblock In \emph{International Conference on Learning Representations
  (ICLR)}, 2019{\natexlab{b}}.

\bibitem[Miller et~al.(2000)Miller, Matsakis, and
  Viola]{Miller-Matsakis-Viola2000}
Miller, E., Matsakis, N., and Viola, P.
\newblock Learning from one example through shared densities on transforms.
\newblock \emph{Conference on Computer Vision and Pattern Recognition (CVPR)},
  2000.

\bibitem[Mishra et~al.(2018)Mishra, Rohaninejad, Chen, and Abbeel]{mishra2018a}
Mishra, N., Rohaninejad, M., Chen, X., and Abbeel, P.
\newblock A simple neural attentive meta-learner.
\newblock In \emph{International Conference on Learning Representations
  (ICLR)}, 2018.

\bibitem[Munkhdalai et~al.(2018)Munkhdalai, Yuan, Mehri, and
  Trischler]{pmlr-v80-munkhdalai18a}
Munkhdalai, T., Yuan, X., Mehri, S., and Trischler, A.
\newblock Rapid adaptation with conditionally shifted neurons.
\newblock In \emph{International Conference on Machine Learning (ICML)}, 2018.

\bibitem[Narasimhan \& Bilmes(2005)Narasimhan and Bilmes]{Narasimhan2005}
Narasimhan, M. and Bilmes, J.
\newblock A submodular-supermodular procedure with applications to
  discriminative structure learning.
\newblock In \emph{Conference on Uncertainty in Artificial Intelligence (UAI)},
  2005.

\bibitem[Oreshkin et~al.(2018)Oreshkin, L{\'o}pez, and
  Lacoste]{oreshkin2018tadam}
Oreshkin, B., L{\'o}pez, P.~R., and Lacoste, A.
\newblock Tadam: Task dependent adaptive metric for improved few-shot learning.
\newblock In \emph{Neural Information Processing Systems (NeurIPS)}, 2018.

\bibitem[Qiao et~al.(2019)Qiao, Shi, Li, Wang, Huang, and Tian]{team}
Qiao, L., Shi, Y., Li, J., Wang, Y., Huang, T., and Tian, Y.
\newblock Transductive episodic-wise adaptive metric for few-shot learning.
\newblock In \emph{International Conference on Computer Vision (ICCV)}, 2019.

\bibitem[Qiao et~al.(2018)Qiao, Liu, Shen, and Yuille]{qiao2018few}
Qiao, S., Liu, C., Shen, W., and Yuille, A.~L.
\newblock Few-shot image recognition by predicting parameters from activations.
\newblock In \emph{Conference on Computer Vision and Pattern Recognition
  (CVPR)}, 2018.

\bibitem[Ravi \& Larochelle(2017)Ravi and Larochelle]{Ravi2017OptimizationAA}
Ravi, S. and Larochelle, H.
\newblock Optimization as a model for few-shot learning.
\newblock In \emph{International Conference on Learning Representations
  (ICLR)}, 2017.

\bibitem[Ren et~al.(2018)Ren, Triantafillou, Ravi, Snell, Swersky, Tenenbaum,
  Larochelle, and Zemel]{ren18fewshotssl}
Ren, M., Triantafillou, E., Ravi, S., Snell, J., Swersky, K., Tenenbaum, J.~B.,
  Larochelle, H., and Zemel, R.~S.
\newblock Meta-learning for semi-supervised few-shot classification.
\newblock In \emph{International Conference on Learning Representations
  {ICLR}}, 2018.

\bibitem[Russakovsky et~al.(2015)Russakovsky, Deng, Su, Krause, Satheesh, Ma,
  Huang, Karpathy, Khosla, Bernstein, Berg, and Fei-Fei]{ILSVRC15}
Russakovsky, O., Deng, J., Su, H., Krause, J., Satheesh, S., Ma, S., Huang, Z.,
  Karpathy, A., Khosla, A., Bernstein, M., Berg, A.~C., and Fei-Fei, L.
\newblock {ImageNet Large Scale Visual Recognition Challenge}.
\newblock \emph{International Journal of Computer Vision (IJCV)}, 115\penalty0
  (3):\penalty0 211--252, 2015.

\bibitem[Rusu et~al.(2019)Rusu, Rao, Sygnowski, Vinyals, Pascanu, Osindero, and
  Hadsell]{rusu2018metalearning}
Rusu, A.~A., Rao, D., Sygnowski, J., Vinyals, O., Pascanu, R., Osindero, S.,
  and Hadsell, R.
\newblock Meta-learning with latent embedding optimization.
\newblock In \emph{International Conference on Learning Representations
  (ICLR)}, 2019.

\bibitem[Shi \& Malik(2000)Shi and Malik]{ShiMalik2000}
Shi, J. and Malik, J.
\newblock Normalized cuts and image segmentation.
\newblock \emph{IEEE Transactions on Pattern Analysis and Machine
  Intelligence}, 22\penalty0 (8):\penalty0 888--905, 2000.

\bibitem[Snell et~al.(2017)Snell, Swersky, and Zemel]{snell2017prototypical}
Snell, J., Swersky, K., and Zemel, R.
\newblock Prototypical networks for few-shot learning.
\newblock In \emph{Neural Information Processing Systems (NeurIPS)}, 2017.

\bibitem[Sun et~al.(2019)Sun, Liu, Chua, and Schiele]{sun2019mtl}
Sun, Q., Liu, Y., Chua, T., and Schiele, B.
\newblock Meta-transfer learning for few-shot learning.
\newblock In \emph{Conference on Computer Vision and Pattern Recognition
  (CVPR)}, June 2019.

\bibitem[Sung et~al.(2018)Sung, Yang, Zhang, Xiang, Torr, and
  Hospedales]{sung2018learning}
Sung, F., Yang, Y., Zhang, L., Xiang, T., Torr, P.~H., and Hospedales, T.~M.
\newblock Learning to compare: Relation network for few-shot learning.
\newblock In \emph{Conference on Computer Vision and Pattern Recognition
  (CVPR)}, 2018.

\bibitem[Szegedy et~al.(2016)Szegedy, Vanhoucke, Ioffe, Shlens, and
  Wojna]{szegedy2016rethinking}
Szegedy, C., Vanhoucke, V., Ioffe, S., Shlens, J., and Wojna, Z.
\newblock Rethinking the inception architecture for computer vision.
\newblock In \emph{Conference on Computer Vision and Pattern Recognition},
  2016.

\bibitem[Tian et~al.(2014)Tian, Gao, Cui, Chen, and Liu]{Tian-AAAI}
Tian, F., Gao, B., Cui, Q., Chen, E., and Liu, T.-Y.
\newblock Learning deep representations for graph clustering.
\newblock In \emph{AAAI Conference on Artificial Intelligence}, 2014.

\bibitem[Vaida(2005)]{vaida2005parameter}
Vaida, F.
\newblock Parameter convergence for em and mm algorithms.
\newblock \emph{Statistica Sinica}, 15:\penalty0 831--840, 2005.

\bibitem[Vaswani et~al.(2017)Vaswani, Shazeer, Parmar, Uszkoreit, Jones, Gomez,
  Kaiser, and Polosukhin]{Vaswani17}
Vaswani, A., Shazeer, N., Parmar, N., Uszkoreit, J., Jones, L., Gomez, A.~N.,
  Kaiser, u., and Polosukhin, I.
\newblock Attention is all you need.
\newblock In \emph{Neural Information Processing Systems (NeurIPS)}, 2017.

\bibitem[Vinyals et~al.(2016)Vinyals, Blundell, Lillicrap, Kavukcuoglu, and
  Wierstra]{Vinyals2016MatchingNF}
Vinyals, O., Blundell, C., Lillicrap, T.~P., Kavukcuoglu, K., and Wierstra, D.
\newblock Matching networks for one shot learning.
\newblock In \emph{Neural Information Processing Systems (NeurIPS)}, 2016.

\bibitem[Von~Luxburg(2007)]{VonLuxburg2007}
Von~Luxburg, U.
\newblock A tutorial on spectral clustering.
\newblock \emph{Statistics and computing}, 17\penalty0 (4):\penalty0 395--416,
  2007.

\bibitem[Wah et~al.(2011)Wah, Branson, Welinder, Perona, and
  Belongie]{wah2011caltech}
Wah, C., Branson, S., Welinder, P., Perona, P., and Belongie, S.
\newblock The caltech-ucsd birds-200-2011 dataset.
\newblock 2011.

\bibitem[Wang \& Carreira-Perpin{\'a}n(2014)Wang and
  Carreira-Perpin{\'a}n]{WangCarreira-Perpinan2014}
Wang, W. and Carreira-Perpin{\'a}n, M.~A.
\newblock The laplacian k-modes algorithm for clustering.
\newblock \emph{Preprint arXiv:1406.3895}, 2014.

\bibitem[Wang et~al.(2019)Wang, Chao, Weinberger, and van~der
  Maaten]{wang2019simpleshot}
Wang, Y., Chao, W.-L., Weinberger, K.~Q., and van~der Maaten, L.
\newblock Simpleshot: Revisiting nearest-neighbor classification for few-shot
  learning.
\newblock \emph{Preprint arXiv:1911.04623}, 2019.

\bibitem[Wertheimer \& Hariharan(2019)Wertheimer and
  Hariharan]{wertheimer2019few}
Wertheimer, D. and Hariharan, B.
\newblock Few-shot learning with localization in realistic settings.
\newblock In \emph{Conference on Computer Vision and Pattern Recognition
  (CVPR)}, 2019.

\bibitem[Weston et~al.(2012)Weston, Ratle, Mobahi, and
  Collobert]{weston2012deep}
Weston, J., Ratle, F., Mobahi, H., and Collobert, R.
\newblock Deep learning via semi-supervised embedding.
\newblock In \emph{Neural networks: Tricks of the trade}, pp.\  639--655.
  Springer, 2012.

\bibitem[Ye et~al.(2020)Ye, Hu, Zhan, and Sha]{ye2020fewshot}
Ye, H.-J., Hu, H., Zhan, D.-C., and Sha, F.
\newblock Few-shot learning via embedding adaptation with set-to-set functions.
\newblock In \emph{Conference on Computer Vision and Pattern Recognition
  (CVPR)}, 2020.

\bibitem[Yuan et~al.(2017)Yuan, Yin, Bai, Feng, and Tai]{Yuan2017}
Yuan, J., Yin, K., Bai, Y., Feng, X., and Tai, X.
\newblock Bregman-proximal augmented lagrangian approach to multiphase image
  segmentation.
\newblock In \emph{Scale Space and Variational Methods in Computer Vision
  (SSVM)}, 2017.

\bibitem[Yuille \& Rangarajan(2001)Yuille and Rangarajan]{Yuille2001}
Yuille, A.~L. and Rangarajan, A.
\newblock The concave-convex procedure {(CCCP)}.
\newblock In \emph{Neural Information Processing Systems (NeurIPS)}, 2001.

\bibitem[Zagoruyko \& Komodakis(2016)Zagoruyko and Komodakis]{WRN}
Zagoruyko, S. and Komodakis, N.
\newblock Wide residual networks.
\newblock In \emph{British Machine Vision Conference (BMVC)}, 2016.

\bibitem[{Zhang} et~al.(2019){Zhang}, {Zhao}, {Ni}, {Xu}, and
  {Yang}]{Variationalfewshot}
{Zhang}, J., {Zhao}, C., {Ni}, B., {Xu}, M., and {Yang}, X.
\newblock Variational few-shot learning.
\newblock In \emph{International Conference on Computer Vision (ICCV)}, 2019.

\bibitem[Zhang et~al.(2007)Zhang, Kwok, and Yeung]{Zhang2007}
Zhang, Z., Kwok, J.~T., and Yeung, D.-Y.
\newblock Surrogate maximization/minimization algorithms and extensions.
\newblock \emph{Machine Learning}, 69:\penalty0 1--33, 2007.

\bibitem[Zhou et~al.(2004)Zhou, Bousquet, Lal, Weston, and
  Sch{\"o}lkopf]{z2004learning}
Zhou, D., Bousquet, O., Lal, T.~N., Weston, J., and Sch{\"o}lkopf, B.
\newblock Learning with local and global consistency.
\newblock In \emph{Neural Information Processing Systems (NeurIPS)}, 2004.

\bibitem[Ziko et~al.(2018)Ziko, Granger, and {Ben Ayed}]{ziko2018scalable}
Ziko, I., Granger, E., and {Ben Ayed}, I.
\newblock Scalable laplacian k-modes.
\newblock In \emph{Neural Information Processing Systems (NeurIPS)}, 2018.

\end{thebibliography}
\end{document}